\documentclass[12pt]{article}

\usepackage[left=2.9cm,right=2.9cm,top=2.2cm,bottom=2.5cm]{geometry} 

\usepackage{amsmath,amssymb,amsfonts,amsthm}
\numberwithin{equation}{section} 

\usepackage{graphicx}
\usepackage{booktabs}
\usepackage{authblk}   
\usepackage{setspace}
\usepackage{natbib}
\usepackage{orcidlink}

\usepackage[normalem]{ulem}

\usepackage{tikz}
\usetikzlibrary{positioning,arrows.meta}

\usepackage{hyperref}

\makeatletter
\renewcommand\hyper@natlinkbreak[2]{#1}
\makeatother
\usepackage[T1]{fontenc}
\usepackage{lmodern}   

\usepackage{microtype}

\usepackage{subcaption}

\usepackage{xcolor}

\hypersetup{
    colorlinks = true,
    linkcolor  = blue,
    citecolor  = blue,
    urlcolor   = blue
}

\newtheoremstyle{myplain}   
  {0.8\baselineskip}        
  {0.5\baselineskip}        
  {\itshape}                
  {}                        
  {\bfseries}               
  {.}                       
  {0.5em}                   
  {}                        

\newtheoremstyle{mydef}
  {0.8\baselineskip}
  {0.5\baselineskip}
  {}                        
  {}
  {\bfseries}
  {.}
  {0.5em}
  {}

\newtheoremstyle{myremark}
  {0.8\baselineskip}
  {0.2\baselineskip}
  {}
  {}
  {\itshape}
  {.}
  {0.5em}
  {}

\theoremstyle{myplain}
\newtheorem{theorem}{Theorem}[section]

\theoremstyle{mydef}

\newtheorem{example}[theorem]{Example}

\theoremstyle{myremark}
\newtheorem{remark}[theorem]{Remark}

\newcommand{\Prob}{\mathbb{P}}

\newcommand{\keywords}[1]{\par\smallskip\noindent\textbf{Keywords:} #1}

\title{\textbf{Expert-Guided Class-Conditional Goodness-of-Fit Scores for Interpretable Classification with Informative Missingness}}

\author[1]{Shahar Cohen\,\orcidlink{0009-0008-9682-2821}\thanks{Corresponding author. Email: \texttt{sc1994@gmail.com}.}}
\author[1]{David M. Steinberg\,\orcidlink{0000-0001-9112-1809}}
\author[2]{Yael Radzyner\,\orcidlink{0009-0002-2026-3185}}
\author[2]{Yochai Ben Horin\,\orcidlink{0000-0001-7046-4511}}

\affil[1]{Department of Statistics and Operations Research, School of Mathematical Sciences, Tel Aviv University, Tel Aviv, Israel}
\affil[2]{Israel NDC, Soreq Nuclear Research Center, Yavne, Israel}

\date{} 

\begin{document}

\maketitle

\begin{abstract}
We study a classification problem with three key challenges: pervasive informative missingness, the integration of partial prior expert knowledge, and the need for interpretable decision rules. We propose a framework that encodes prior knowledge through an expert-guided class-conditional model for one or more classes, and uses this model to construct a small set of interpretable goodness-of-fit features. The features quantify how well the observed data agree with the expert model, isolating the contributions of different aspects of the data, including both observed and missing components. These features are combined with a few transparent auxiliary summaries in a simple discriminative classifier, resulting in a decision rule that is easy to inspect and justify. We develop and apply the framework in the context of seismic monitoring used to assess compliance with the Comprehensive Nuclear-Test-Ban Treaty. We show that the method has strong potential as a transparent screening tool, reducing workload for expert analysts. A simulation designed to isolate the contribution of the proposed framework shows that this interpretable expert-guided method can even outperform strong standard machine-learning classifiers, particularly when training samples are small.
\end{abstract}


\newpage



\section{Introduction}

Machine learning (ML) classification techniques are increasingly used to automate tasks traditionally performed by human experts. A key goal in such settings is to replicate the specialized judgment of human experts. However, black-box models often lack transparency, making them unsuitable when users need to understand the rationale behind algorithmic decisions \citep{caruana2015intelligible, rudin2019stop}. On the other hand, fully specifying expert decision rules is typically infeasible: this is precisely why ML is employed in the first place. The growing reliance on machine learning in high-stakes settings has sharpened the demand for interpretable models that are transparent, easy to justify, and grounded in meaningful features \citep{rudin2019stop, Nutovitz2023XAI}.

This challenge is further amplified when expert decisions rely not only on observed data, but also on informative missing data, where insights are drawn from what is \textit{not} present in the data. Missing data is often a central obstacle in machine learning and is especially acute in settings where standard assumptions (e.g., missing at random, MAR, in the sense of \cite{little2019statistical}) are violated. In many applications, missingness is not merely present, it is frequent, structural and informative: the fact that a value is missing can encode information about mechanisms, processes, or selection effects, so treating missing entries as mere noise, or replacing them by imputation, can distort the learning problem \citep{lipton2016directly,che2018recurrent,zhou2024comprehensive,lin2025addressing}. 

A constructive response to the above challenges is to integrate domain expertise directly into the classifier’s design \citep{karpatne2017theory,deng2020integrating,karniadakis2021,vonrueden2023informed}. Integrating expert knowledge into data-driven models is an open research challenge. We often have domain understanding that could benefit the learning process, but translating that understanding into usable constraints, features, priors, or model components that fit modern classification pipelines is still nontrivial \citep{deng2020integrating,vonrueden2023informed}.

This work is motivated by a case study that brings these three threads together: (i) addressing classification under pervasive structural missingness, (ii) embedding expert knowledge into classification, and (iii) designing classifiers that are both transparent and easy to justify, through the construction of a small number of meaningful, interpretable features. The methodology we develop for our motivating application tackles these three challenges jointly and can also be applied to each one in isolation.

The method relies on constructing domain-informed, interpretable features, each designed to capture a meaningful, expert-driven aspect of the data.  Some of the features are specifically constructed to account for informative missing data; however, the methodology is general and it can incorporate a wide range of expert knowledge, not only information related to missingness. Once these features have been defined, we can train a relatively simple and interpretable classifier, such as logistic regression, that enables understanding of how the expert concepts affect the classifier. Alternatively, the expert features can be used as additional inputs to less transparent classification procedures when explainability is not a major concern.

Our primary example comes from seismic monitoring of the Comprehensive Nuclear-Test-Ban Treaty (CTBT), which is used to detect underground explosions that violate the treaty \citep{wu2021ctbt,richards2009monitoring}. An important hurdle in the monitoring is to determine whether a candidate seismic event corresponds to a genuine event (e.g., an earthquake or explosion), which we formulate as a binary classification problem. The automated processing pipeline contains many errors. As a result, experts must review events one by one to determine which events are valid. This manual review process is costly and time-consuming \citep{draelos2012false,benhorin2013scoring}. Crucially, it relies on prior knowledge, including physics-based models of wave propagation, that guides the expert decision.

The key insight is that when experts evaluate a proposed event (which includes both the observations from multiple sensors and estimated parameters, such as event location), their main task is to assess whether the observations are consistent with what theoretical physics would predict for an event with those attributes. They are effectively performing a physics-based goodness-of-fit check, asking whether the pattern of observed and missing features is compatible with the physical model. This type of decision-making is not limited to physics models. For example, when physicians make a diagnosis, they ask whether the entire pattern of symptoms and test results is consistent with a proposed pathology. Often, as in our application, symptoms that are \emph{not} present convey valuable information.

To formalize this intuition, our method assumes access to an expert-guided model describing the behavior of the observed data conditional on instance-specific attributes, for at least one class of interest. In seismic monitoring, these attributes include event location, origin time, and magnitude, and the model specifies which stations are expected to detect the event and what measurements are expected when they do. More generally, the reference model may derive from a scientific or mechanistic theory, a hybrid of theory and data-driven estimation, or a learned conditional model whose structure is informed by domain expertise. We use this model to construct a small set of interpretable features by quantifying how well each candidate instance agrees with the model. Decomposing the agreement score results in features that indicate not only whether an instance is inconsistent with a proposed class label, but also how it is inconsistent. Because missingness is incorporated directly into the reference model, the absence of an observation can itself provide evidence for or against a class.

We presented the basic ingredients of our proposal for the seismic monitoring application in our earlier work \citep{cohen2026improved}. That paper included a detailed empirical evaluation of the resulting classifier. Our goal in this paper is to develop the broader statistical framework underlying the approach, allowing us to formulate the methodology in a general setting beyond the specific seismic monitoring application. We also investigate its properties through the data analysis and a dedicated simulation study.

The remainder of the paper is organized as follows. Section~\ref{sec:motivation} presents the motivating example. Section~\ref{sec:setup} introduces the notation, formal problem setup, and related work. Section~\ref{sec:methodology} presents the methodology in detail. Section~\ref{sec:data} applies the ideas to our seismic monitoring problem. Section~\ref{sec:simulations} gives results from a simulation constructed to test the methodology. Conclusions are presented in Section~\ref{sec:conclus}.

\section{Motivating Example: Monitoring the CTBT}
\label{sec:motivation}

In seismic monitoring, a central task is the construction of a reliable bulletin of seismic events \citep{draelos2015new}. We consider this problem in the context of the Comprehensive Nuclear-Test-Ban Treaty (CTBT), where such bulletins  are especially important because seismic data are the primary tool to detect and identify underground nuclear explosions \citep{wu2021ctbt}. The initial product is an automatic bulletin that is timely, but not reliable: in our dataset, more than 58\% of automatically generated events were later rejected after expert review (see Section~\ref{sec:data}).

Seismic monitoring relies on data collected from a network of stations that continuously record ground motion. A seismic event (e.g., an earthquake or a large explosion) generates ground motion that can trigger a detection at a station. When several stations report detections within a short time window, this suggests that an event has occurred. The data recorded at these stations are then used to estimate important characteristics such as its location and magnitude. The IDC applies an algorithm that associates detections into candidate ``events'' using physical wave-propagation models and heuristics \citep{tibi2019iterative,draelos2015new,mcbrearty2019association}.

Human experts review the automatically generated bulletins, using their expertise to rule out many associations as ``false events''. One important source of information is \emph{non-detecting stations}. When forming events, the automatic association process considers only the detecting stations, ignoring non-detecting stations \citep{arora2013netvisa} as if the data from them was MAR. The resulting event is thus reasonably consistent with the data from the detecting stations; however, the physical model may imply that a non-detecting station should have detected the event. Such non-detects are important indicators to human analysts of a false event. The CTBT setting is especially challenging because its sparse global network makes non-detections common and can lead automated pipelines to combine unrelated distant detections into false events \citep{gaebler2021performance,hutton2006southern}.

The review process, in which many events are rejected by the experts, yields a much more reliable bulletin. Our goal is to integrate this expert knowledge into the automatic system, so that the initial bulletin already incorporates this expertise and eliminates most of the false events. We formulate the problem as a classification task. We have a dataset of events from a low-quality bulletin, and we also have a reliable expert-reviewed bulletin. Therefore, we have a label indicating whether a candidate event from the automatic bulletin also appears in the expert-reviewed bulletin.

However, the situation has two unique aspects. First, because the bulletins are used to monitor CTBT compliance, decisions must be explainable and grounded in the physics-based reasoning experts already use. Second, missingness is highly informative: for weak events in a sparse global network, most stations record no signal, and the resulting non-detection pattern provides key evidence about whether a proposed event is physically plausible.

\section{Setting, Assumptions, and Related Work}
\label{sec:setup}
The general problem we consider is supervised classification, in which $(X,Y)$ is a pair of random variables, $X$ takes values in a feature space $\mathcal{X}$ and the class label $Y$ takes values in $\mathcal{Y}=\{1,\dots,K\}$. We restrict attention to the binary case $K=2$, although extension to the multiclass setting is immediate. 

\subsection{Sensor-based missingness in the feature space}

In our setting, each example aggregates data from $S$ distinct sensors. We take the (latent) complete feature object to be a block-structured vector
\[
X = \bigl(X^{(1)},\dots,X^{(S)}\bigr) \in \mathcal{X}, 
\qquad 
\mathcal{X} := \prod_{s=1}^S \mathbb{R}^q = (\mathbb{R}^q)^S,
\qquad
X^{(s)} \in \mathbb{R}^q,
\]
where $X^{(s)}$ denotes the $q$-dimensional component contributed by sensor $s$. One may allow different block dimensions across sensors (i.e., $X^{(s)}\in\mathbb{R}^{q_s}$), but for simplicity, and in accordance with our seismological example, we assume a common dimension $q$. In practice, it is also possible that the number of sensors varies from one example to another (and in fact that is the case in our application, as seismic stations have periods of down-time when they are not active). Generalization to that setting is trivial. To simplify, we present only the case of constant $S$.

Missingness arises at the \emph{sensor level}: if sensor $s$ is missing, then the entire block $X^{(s)}$ is unobserved. We associate to $X$ a sensor-missingness indicator
$
D = (D^{(1)},\dots,D^{(S)}) \in \{0,1\}^S,
$
where, for each sensor $s\in\{1,\dots,S\}$,
\[
D^{(s)} =
\begin{cases}
1, & \text{if sensor $s$ is observed},\\
0, & \text{if sensor $s$ is missing}.
\end{cases}
\]
Let $\widetilde{X} = \bigl(\widetilde{X}^{(1)},\dots,\widetilde{X}^{(S)}\bigr)$ denote the corresponding partially observed object, with
\[
\widetilde{X}^{(s)} =
\begin{cases}
X^{(s)}, & \text{if } D^{(s)} = 1,\\
\texttt{NA}_q, & \text{if } D^{(s)} = 0,
\end{cases}
\]
where $\texttt{NA}_q$ denotes a $q$-dimensional missing-value placeholder. The data for the \(i\)th example are $
Z_i = (\widetilde{X}_i, D_i, Y_i)$,
and the sample is
$
\mathcal{D}_n = \{Z_i\}_{i=1}^n. 
$ Sensor-level missingness is adopted for convenience, coordinate-wise missing features can be  included as the special case \(q=1\). We do not assume MAR. Let
$
X_{\mathrm{obs}} := \{X^{(s)} : D^{(s)}=1\}
$
denote the collection of observed sensor blocks. Thus, we allow
$
P(D \mid X,Y) \not\equiv P(D \mid X_{\mathrm{obs}},Y).
$
Moreover, the missingness pattern may itself be informative for classification, so we allow
$
P(Y \mid X_{\mathrm{obs}},D) \not\equiv P(Y \mid X_{\mathrm{obs}}).
$

\subsection{Expert-guided class-conditional models}
\label{sec:expertguided}

We assume access to expert/scientific/prior knowledge in the form of a (possibly partial) semi-generative \emph{expert-guided model} for the observed data \((\widetilde{X},D)\). 
Formally, for at least one class label \(y\in\mathcal{Y}\), we posit a conditional family
\[
(\widetilde{X}_i,D_i)\mid Y_i=y\ \sim\ f_{y}(\cdot ; \theta_i),
\]
where \(\theta_i\in \Theta_y\) denotes instance-specific parameters or contextual variables, and \(f_{y}(\cdot; \theta)\) is a model for the distribution of \((\widetilde{X},D)\) in class \(y\). This assumption encodes prior knowledge as a distributional statement: conditional on \(Y=y\), and for a fixed (potentially unobserved) parameter vector \(\theta\), the observed variables \((\widetilde{X},D)\) are generated according to \(f_{y}(\cdot; \theta)\). We consider two interpretations of \(\theta\):
\begin{enumerate}
    \item \textbf{Instance-specific latent parameters.} 
    \(\theta\) is an unobserved, instance-dependent latent variable that governs the generation of \((\widetilde{X},D)\) conditional on \(Y=y\). 
    In this case, \(f_{y}(\cdot ; \theta)\) is a semi-generative model indexed by a latent state \(\theta\).

    \item \textbf{Observed context.} 
    \(\theta\) is observed for each instance and is treated as part of the available data. The feature set is partitioned into \emph{response} (or \emph{explained}) variables \((\widetilde{X},D)\) and \emph{conditioning} (or \emph{context}) variables \(\theta\). In this case, the observed data for inference are \((\theta_i,\widetilde{X}_i,D_i,Y_i)_{i=1}^n\).
\end{enumerate}

\begin{example}
    In our seismic monitoring setting, interpretation (1) is natural: the latent event attributes (e.g., location, magnitude, source mechanism) play the role of $\theta$, and the physics of wave propagation induce a distribution for which stations are likely to record a signal and how the recorded waveforms/features should look. Interpretation~(2) may arise, for example, in medical applications where patient context (e.g., age, sex, comorbidities, baseline measurements) is fully observed. In that case, expert knowledge can specify how “healthy” or “pathological” test results should be distributed conditional on context, including a model of the testing process (i.e., when and for whom tests are ordered) and thus of missingness as well. Interpretation (2) arises in many applications. One example is medical imaging, where anatomical measurements can be predictive of observed data. In positron emission tomography (PET), the model for observed detector-pair counts, which is derived from photon-propagation physics, exploits an accompanying CT scan of the patient's anatomy, which is used to compute an attenuation map \citep{schulz2011automatic}.
\end{example}

\begin{remark}
Importantly, we do not require an expert-guided model for every class label. This is clearly relevant in the seismic monitoring setting: station data for genuine seismic events should conform to the scientifically grounded model for wave propagation. In contrast, the class of invalid events produced by automatic association is heterogeneous and pipeline-dependent, and is not naturally described by a generative model. Our framework therefore allows us to exploit a class-conditional model where expert knowledge is available (e.g., the ``valid event'' class) without requiring an equally detailed generative specification for all classes.
\end{remark}

\paragraph{How the model is obtained.}
We treat \(f_y(\cdot;\theta)\) as given, but the methodology applies whenever such a conditional model is available, including the following cases:
\begin{enumerate}
    \item \textbf{Complete scientific knowledge.} Models are derived entirely from scientific or domain knowledge.

    \item \textbf{Hybrid (partially specified + learned).}
    Some components are fixed by expert knowledge, while others are learned from data. For example, one may posit
    \[
    \theta_i = g(X_i) + \varepsilon_i,
    \qquad \mathbb{E}[\varepsilon_i]=0,
    \]
    where \(g\) is expert-specified and \(\varepsilon_i\) captures discrepancy and stochastic variation. After fitting the model, the residuals can be used to estimate the distribution of \(\varepsilon\), thereby completing the model specification.
    
    \item \textbf{Data-driven with optional expert guidance.}
    The model is learned from data, including the conditional generative structure
    $
    f(\widetilde{X},D \mid \theta, Y=y).
    $
    Experts may constrain the learning problem, for example by specifying or restricting the context-response partition.
\end{enumerate}

The seismological example corresponds to case (2), and we describe how we construct the models in Section~\ref{sec:data}. 

\begin{remark}
If any component of \(f_y(\cdot;\theta)\) is learned from data, model fitting should be separated from the methodology introduced below to avoid bias. Standard separation methods include sample splitting or an external training set.
\end{remark}

\subsection{Objectives}

Given the structured observations, our objective is to obtain a classifier
\[
h:\widetilde{\mathcal{X}}\times\{0,1\}^S \to \mathcal{Y},
\qquad
h\bigl(\widetilde{X},D\bigr)\approx Y,
\]
where \(\widetilde{\mathcal{X}}:=\prod_{s=1}^S\bigl(\mathbb{R}^q\cup\{\texttt{NA}_q\}\bigr)\) denotes the space of partially observed sensor blocks. We seek a decision rule that: (i) classifies well under standard performance criteria; (ii) is transparent, in the sense that it can be expressed through a small set of domain-interpretable features that allow experts to inspect and understand individual decisions; (iii) treats sensor availability and non-detections as potentially informative by explicitly leveraging the missingness pattern rather than defaulting to imputation or ignoring missing values; and (iv) incorporates expert domain knowledge about how the observed measurements \(\widetilde{X}\) and missingness pattern \(D\) behave under different class labels, so that decisions reflect scientifically meaningful constraints and explanations.

\subsection{Related work}

A comprehensive review of all the challenges described above and methods for addressing them is beyond the scope of this paper. We therefore focus on the work most closely related to our approach and refer readers to surveys on (a) missing-data methods for prediction and classification \citep{saar2007handling,emmanuel2021survey,miao2023experimental}, (b) explainable and interpretable machine learning \citep{dwivedi2023xai,allen2024interpretable}, and (c) informed or knowledge-guided machine learning \citep{karniadakis2021,willard2022integrating,vonrueden2023informed}. Additionally, the fact that only a subset of class labels is modeled (the valid-event class in our application) is reminiscent of novelty- and one-class detection methods, which learn a description of normal data and identify departures from it \citep{scholkopf2001estimating,tax2004support,pimentel2014review}. Our procedure differs in that it uses labeled invalid events and a supervised final classifier.

A classical distinction is between \emph{generative} and \emph{discriminative} classifiers \citep{ng2002discriminative,jebara2004machine}. Generative classifiers model the joint distribution $p(x,y)$, often by factorizing it as $p(y)\,p(x\mid y)$, and make predictions via Bayes' rule. Discriminative approaches instead model $p(y\mid x)$ directly. Hybrid methods seek to combine these paradigms in order to capture their complementary advantages \citep{jaakkola1998exploiting,raina2003classification,kuleshov2017deep,hayashi2024hybrid}. A canonical example is the \emph{Fisher kernel} \citep{jaakkola1998exploiting}, where a fitted generative model is used to map each instance to its \emph{Fisher score} (the gradient of the log-likelihood with respect to model parameters), yielding a discriminative kernel. Generally, generative models can serve as representation learners whose outputs, such as likelihoods, or parameter estimates, are then used by discriminative predictors \citep{brodersen2011generative,perina2012free}. Our approach follows this general philosophy.

A closely related line of work constructs \emph{likelihood-space} representations, in which each sample is mapped, via a family of generative models with one model per class label, to a vector of (log-)likelihoods, and classification is then performed discriminatively in that space \citep{bicego2009component}. This approach is related to our use of likelihood-based model-fit scores as features. However, it assumes a fixed global collection of fully specified generative models. We relax this framework in two ways. First, we allow $f_y(\cdot;\theta)$ to vary across instances through an instance-specific parameter $\theta$, rather than assuming a single global generative model for each class. This extension also accommodates settings in which $\theta$ is observed and induces a partition of the feature space. Second, we do not require a generative model for every class label.

Hybrid ideas have also been used in learning problems with missing data \citep{ipsen2022missing,pingi2024joint}, where the generative component is used either to infer a distribution over the missing entries \citep{mattei2019miwae,ipsen2022missing} or to learn a low-dimensional latent representation from partially observed inputs \citep{ma2020vaem,gong2021vsae}. The general motivation is that the generative part can exploit structure in incomplete covariates, while the discriminative part remains targeted to the prediction task \citep{ipsen2022missing}. One recent work, by \citet{belenguerllorens2026interpretable}, explicitly discusses using such methods for handling missing values, with an emphasis on interpretability, in a multimodal setting, where the model imputes missing modalities and learns an interpretable latent representation for prediction. 

Within the interpretability literature, our approach belongs to the broader class of \emph{intrinsically interpretable} methods, rather than methods that rely on \emph{post hoc} explanations of black-box predictors \citep{allen2024interpretable,dimarino2025antehoc}. In such approaches, interpretability is built directly into the modeling pipeline, by constructing a meaningful intermediate representation and then applying a transparent decision rule \citep{koh2020concept,alvarezmellis2018self}. Our method follows this perspective, which is often argued to be preferable in high-stakes settings when competitive predictive performance is achievable \citep{rudin2019stop}.

\section{Methodology}
\label{sec:methodology}

We now present the general methodology underlying the proposed approach. As illustrated in Figure~\ref{fig:pipeline_overview}, the proposed pipeline first estimates instance-level parameters under each expert-guided class-conditional model. It then uses the fitted models to construct an augmented representation, consisting of interpretable, class-specific, model-fit scores and additional auxiliary features. Finally, a transparent classifier is trained on this representation. At a high level, expert guidance enters the pipeline in three distinct ways:
\begin{enumerate}
    \item \textbf{Class-conditional models:} for one or more labels \(y\) in a modeled subset \(\mathcal M \subseteq \mathcal Y\), we specify class-conditional model families $f_y(\cdot;\theta)$, in any of the ways described in Section \ref{sec:expertguided}.
    \item \textbf{Score decomposition:} In addition to specifying the probabilistic parametrized model, expert knowledge is used to identify an interpretable decomposition of the induced model-fit score (e.g., log-likelihood) into components aligned with meaningful parts of the data-generating process. Each component captures a distinct aspect of fit (for example, contributions from detection patterns, or observed signal values), which improves both transparency and diagnostic interpretability. This can be viewed as a refinement of Point~1.
    \item \textbf{Auxiliary feature selection:} expert knowledge guides the construction and selection of a small set of transparent auxiliary features, based on both the original data and the instance-level $\theta_i$. 
\end{enumerate}

\begin{figure}[!htbp]
\centering
\begin{tikzpicture}[
    >=Latex,
    box/.style={
        draw,
        rounded corners,
        align=center,
        inner sep=3pt,
        minimum height=8mm,
        text width=4.15cm
    },
    data/.style={box, fill=blue!6},
    expert/.style={box, fill=green!8},
    fitstep/.style={box, fill=orange!10},
    arr/.style={->, thick}
]

\node[data]    (A) at ( -3.0,  0.0) {\textbf{Data.} \\Raw observed (multi-sensor) data \\$(\widetilde{X}_i,D_i)$};
\node[expert]  (B) at ( 3.0, 0.0) {\textbf{Models.} \\ Expert-guided\\Class-conditional model(s)\\$\{f_y(\cdot;\theta)\}_{y\in\mathcal M}$};

\node[expert]  (C) at (-5.0, -4.0) {\textbf{Feature selection.} Expert-guided\\ transparent summaries to include};
\node[fitstep] (D) at ( 0.0, -3.5) {\textbf{Instance-level parameters.}\\ (via fitting) $\{\hat{\theta}_i^{(y)}\}_{y\in\mathcal M}$};
\node[expert]  (E) at ( 5.0, -4.0) {\textbf{Decomposition.} Expert-guided\\Likelihood decomposition\\to interpretable score components};

\node[data]    (F) at (-2.5, -7.5) {\textbf{Auxiliary features.}\\Data-derived auxiliary\\transparent features / summaries\\$a_i$ (counts, diagnostics, etc.)};
\node[data]    (G) at ( 2.5, -7.5) {\textbf{Model-fit scores.} \\ Total and / or decomposed scores\\$\{\ell_{i}^{(y)}\}_{y\in\mathcal M}$};

\node[fitstep] (H) at ( 0.0, -11.5) {\textbf{Classification. }Fit a simple interpretable classifier\\on the augmented representation\\$\phi_i=[\{\ell_{i}^{(y)}\}_{y\in\mathcal M},a_i]$};

\draw[arr] (A.south) -- (C.north);
\draw[arr] (A.south) -- (E.north);
\draw[arr] (A.south) -- (D.north);

\draw[arr] (B.south) -- (C.north);
\draw[arr] (B.south) -- (E.north);
\draw[arr] (B.south) -- (D.north);

\draw[arr] (D) -- (E);
\draw[arr] (D) -- (C);

\draw[arr] (C) -- (F);

\draw[arr] (E) -- (G);

\draw[arr] (F.south) -- (H.north);
\draw[arr] (G.south) -- (H.north);

\end{tikzpicture}
\caption{
Overview of the proposed pipeline.
}
\label{fig:pipeline_overview}
\end{figure}

\subsection{Step 1: Class-specific estimation}
\label{subsec:step1_fitting}

The following formulation covers the cases in which the context \(\theta\) is not directly observed. If \(\theta\) is fully observed, simply set \(\hat{\theta}_i^{(y)}=\theta_i\) and use the rest of the pipeline.

For each modeled class \(y\in\mathcal M\), define a class-specific estimation criterion
\[
Q_y(\theta; z), \qquad \theta\in\Theta_y,
\]
chosen so that larger values correspond to better fit of the observed instance \(z\) under class label \(y\).
In the likelihood-based case,
$
Q_y(\theta; z)=\log f_\theta(z|y).
$

More generally, $Q_y$ may be augmented with a regularization term or, in a Bayesian formulation, with a log-prior. For each instance \(i\) and class hypothesis \(y\in\mathcal M\), compute the estimator
\begin{equation}
\hat{\theta}_i^{(y)} \in \arg\max_{\theta\in\Theta_y} Q_y(\theta; z_i).
\label{eq:instance_fit_general}
\end{equation}

A key point is that the estimation in \eqref{eq:instance_fit_general} is performed separately for each instance \(i\) and each class hypothesis \(y\in\mathcal M\). This is essential for deployment: when a new instance is presented, we do not know in advance whether \(y\) is the correct label. Consequently, we also do not know whether the expert-guided class-conditional model for \(y\) provides an adequate description of that instance, or even whether a meaningful underlying state \(\theta\) exists for that instance. In our motivating example, for the class of false events, there are no physical event attributes to estimate at all. 

\subsection{Step 2: Feature engineering}
\label{subsec:step2_feature_engineering}

Step~2 constructs features that represent each instance by \emph{how well it fits the expert-guided class label model}. This step is inherently guided by expert knowledge and depends substantially on domain-specific modeling assumptions. We present here the general methodology and its specific implementation in our seismic monitoring application. 

\subsubsection{Model-fit score features from a multi-source likelihood decomposition}

For each instance \(i\) and modeled class \(y\in\mathcal M\), we use the fitted state \(\hat{\theta}_i^{(y)}\) from Step~1 and compute features that quantify compatibility with class label \(y\), with the primary model-fit score being the fitted log-likelihood
\begin{equation}
\ell_i^{(y)} = \log f_y\bigl(z_i ;{\hat{\theta}_i^{(y)}}\bigr).
\label{eq:step2_total_score}
\end{equation}

\paragraph{Multi-sensor mixed-data factorization.}
To make the construction explicit, suppose the class-conditional model provides, for each sensor \(s\in\{1,\dots,S\}\),
\begin{itemize}
    \item a sensor-observation (detection) probability
    \[
    \pi_{y,s}(\theta) := \Prob\!\bigl(D^{(s)}=1 \mid Y=y,\Theta=\theta\bigr),
    \]
    \item and a conditional density (or probability) for the observations from sensor $s$, 
    \[
    g_{y,s}\!\bigl(x^{(s)};\theta\bigr)
    := f\!\bigl(X^{(s)}=x^{(s)} \mid D^{(s)}=1, Y=y,\Theta=\theta\bigr).
    \]
\end{itemize}
(When \(X^{(s)}\) is mixed discrete/continuous, \(g_{y,s}\) is understood as the appropriate mixed conditional law.) Assuming conditional independence across sources (given \(y\) and \(\theta\)), the mixed density factorizes as
\begin{equation}
\begin{aligned}
f_\theta(z_i|y)
&=
\underbrace{\prod_{s:\,D_i^{(s)}=1}\pi_{y,s}(\theta)}_{\text{detection probability}}
\;\cdot\;
\underbrace{\prod_{s:\,D_i^{(s)}=0}\bigl[1-\pi_{y,s}(\theta)\bigr]}_{\text{non-detection probability}}
\;\cdot\;
\underbrace{\prod_{s:\,D_i^{(s)}=1} g_{y,s}\!\bigl(x_i^{(s)};\theta\bigr)}_{\text{summaries}},
\end{aligned}
\label{eq:source_likelihood_factorization_general}
\end{equation}
where the \(g_{y,s}\)-term appears only when \(D_i^{(s)}=1\) (i.e., when source \(s\) is observed for instance \(i\)). This yields a tractable likelihood and a transparent decomposition into detection, non-detection, and observed-value score components.

\begin{remark} 
The simplifying assumption of conditional independence across sensors may not hold exactly. Residual dependence may remain due to unmodeled common factors or misspecification. Still, conditional independence is a reasonable working assumption when the expert model, including $\theta_i^{(y)}$, captures the primary effects of shared drivers of the sensors. In our seismological setting, inter-station dependence is largely explained
by event attributes and station location. 
\end{remark}

Evaluating \eqref{eq:source_likelihood_factorization_general} at \(\hat{\theta}_i^{(y)}\) and taking logs yields an additive decomposition that naturally defines interpretable score components:

\begin{align}
\ell_{i,\mathrm{det}}^{(y)}
&:=
\sum_{s=1}^S D_i^{(s)} \log \pi_{y,s}\!\bigl(\hat{\theta}_i^{(y)}\bigr),
\label{eq:det_component}
\\
\ell_{i,\mathrm{nondet}}^{(y)}
&:=
\sum_{s=1}^S (1-D_i^{(s)}) \log \!\Bigl(1-\pi_{y,s}\!\bigl(\hat{\theta}_i^{(y)}\bigr)\Bigr),
\label{eq:nondet_component}
\\
\ell_{i,\mathrm{obs}}^{(y)}
&:=
\sum_{s=1}^S D_i^{(s)} \log g_{y,s}\!\bigl(x_i^{(s)};\hat{\theta}_i^{(y)}\bigr),
\label{eq:obs_component}
\end{align}
so that
$
\ell_i^{(y)}
=
\ell_{i,\mathrm{det}}^{(y)}
+
\ell_{i,\mathrm{nondet}}^{(y)}
+
\ell_{i,\mathrm{obs}}^{(y)}.
$

\begin{remark}
Non-detections enter the score directly through \(\ell_{i,\mathrm{nondet}}^{(y)}\), so missing observations are treated as informative evidence rather than ignored or imputed. In the CTBT setting, this matches the analyst's question of whether a station should reasonably have detected an event with fitted properties \(\hat{\theta}_i^{(y)}\).
\end{remark}

Because instances may involve different numbers of observed or active sources, we use normalized score features, such as average log-likelihood per source or average observed-value score per detected source. The denominator should match the intended interpretation of the component. The same decomposition can be refined when useful. For example, if a sensor provides several measurements, the observed-value score may be split into separate components, such as arrival-time fit and amplitude fit. Similarly, the total score may be decomposed into sensor-level contributions, which can be used diagnostically to identify unusually poor-fitting sensors or to summarize fit within expert-defined sensor groups.

Taken together, the construction above yields a family of interpretable model-fit features for each class hypothesis \(y\in\mathcal M\). These include a total fitted score, scientifically meaningful component scores (e.g., detection, non-detection, and observed-value terms) and, when useful, finer decompositions and sensor-level diagnostics. 

\subsubsection{Additional transparent auxiliary summaries and final feature vector}

In addition to the model-fit score features, we use expert guidance to construct a small set of transparent auxiliary summaries \(a_i \in \mathbb{R}^p\). These summaries capture simple aspects of the observed multi-source pattern, properties of the fitted states \(\{\hat{\theta}_i^{(y)}\}_{y\in\mathcal M}\), and domain-informed quantities obtained by combining fitted states with expert-specified structural relationships or model-based transformations. Formally, one may write
\[
a_i = A\!\left(
z_i,\{\hat{\theta}_i^{(y)}\}_{y\in\mathcal M}
\right),
\]
where \(A(\cdot)\) is an expert-guided feature-construction map. The final augmented representation used in Step~3 is
\[
\phi_i
=
\bigl[\{u_i^{(y)}\}_{y\in\mathcal M},\, a_i\bigr],
\]
where \(u_i^{(y)}\) denotes the chosen score components for class label \(y\).

\begin{remark}
An important consequence of the feature-construction step is that it supports \emph{decision-level exclusion} of variables that may be predictive but are undesirable as explicit screening criteria. In our CTBT application, location-related features may help predict whether an event is valid, but directly screening events by location may conflict with the goals of the CTBT (for example, by creating an exploitable criterion for bad actors seeking to avoid detection). Our framework allows such variables to be used only indirectly, through the expert-guided modeling and score construction, while excluding them as explicit inputs to the final classifier. This is useful when certain variables are scientifically relevant for modeling expected observations, but should not appear directly in the final decision rule.
\end{remark}

\subsection{Step 3: Classification on the augmented representation}
\label{subsec:step3_classifier}

After constructing the augmented representation \(\phi_i\), we fit a transparent discriminative classifier
$
h:\Phi \to \mathcal Y,
\ \Phi \subseteq \mathbb{R}^{p_{\text{aug}}}.
$
When interpretability is a direct goal, the classifier is chosen to preserve interpretability of decisions in terms of the score features and auxiliary summaries. Simple examples include logistic (or multinomial) regression and small decision trees. In settings where interpretability is secondary, the augmented representation can be used as extra features in any classifier.

\section{Data Analysis}
\label{sec:data}
\subsection{Dataset}

The working dataset was constructed using historical data from the International Data Centre (IDC) of the Comprehensive Nuclear-Test-Ban Treaty Organization (CTBTO). We used the Standard Event List 1 (SEL1) bulletin, the first bulletin in the IDC processing pipeline, as the primary data source for evaluation and improvement. Because the bulletin is extensive, the dataset was created through careful filtering and selection. We summarize here key aspects of the dataset generation process. Further details are provided in the Supplementary Material.

The refined SEL1 dataset covers the period from October 2013 to November 2023, with a total of 587,407 events. Each event is characterized by five parameters: $\theta_{\mathrm{lat}}$, $\theta_{\mathrm{lon}}$, $\theta_{\mathrm{depth}}$, $\theta_{\mathrm{time}}$, and $\theta_{\mathrm{mag}}$, representing the event’s latitude, longitude, depth, origin time, and body wave magnitude, respectively.

Data were available from 45 primary seismic stations that transmit real-time seismograms to the IDC. Each station either recorded a detection or a non-detection (missing), depending on whether an arrival of a seismic wave at that station was associated with the event. Non-detections are attributed to the signal being too weak to exceed the background noise level at the station. An additional source of missingness could arise from station inactivity, for example due to technical failures or maintenance, and was therefore treated as independent of the event attributes (MCAR). For detecting stations, we considered two standard summaries of the observed seismic signal: (i) $\log(\mathrm{amplitude}/\mathrm{period})$ (log(a/T)), the logarithm of the wave amplitude divided by its period, and (ii) arrival time. These quantities are important observables that depend strongly on the characteristics of the event. 

Labels indicating whether events reported in SEL1 were valid or invalid were derived using the Late Event Bulletin (LEB), a meticulously reviewed IDC bulletin in which events are manually validated by experts. Labels were assigned according to a standard seismological matching criterion: spatial distance less than 5°, origin time within ±60 seconds, and matching arrivals from at least two stations. An event in SEL1 was considered valid if it had a unique corresponding matching event in the LEB. Following these rules, 41.63\% of the SEL1 events were declared valid.

Data collected before November 1, 2022 were used for training and validation, while data from November 1, 2022 to November 1, 2023 were reserved for testing (61,658 events). Using a full year of data for testing ensures that the results will be free from seasonal effects.

\subsection{Expert-guided models, feature engineering and classification}

The expert-guided models were constructed using a hybrid approach. The physics of wave propagation provides a partially specified model and structural constraints, leaving certain components to be estimated from data. These components were estimated using the training and validation data from the LEB excluding the test period. For the valid event class, the general structure of the expert-guided model is similar to the factorization given in Equation~\ref{eq:source_likelihood_factorization_general}, and is given by
\begin{align}
L\!\left(\theta_i; \{d_i^{(s)},x_i^{(s)}\}_{s=1}^{S}\right)
&=
\prod_{\substack{d_i^{(s)}=1}}
f_{X_{i,\mathrm{time}}^{(s)} \mid D_i^{(s)}=1;\theta_i}
\!\left(x_{i,\mathrm{time}}^{(s)}\right)
\cdot
\prod_{\substack{d_i^{(s)}=1}}
f_{X_{i,\mathrm{log(a/T)}}^{(s)} \mid D_i^{(s)}=1;\theta_i}
\!\left(x_{i,\mathrm{log(a/T)}}^{(s)}\right)
\notag\\
&\quad\cdot
\prod_{\substack{d_i^{(s)}=1}} \pi_s(\theta_i)
\cdot
\prod_{\substack{d_i^{(s)}=0}} \left(1-\pi_s(\theta_i)\right).
\label{eq:likelihood_bottom_up_split}
\end{align}

Here, $\theta_i$ denotes the latent attributes of event $i$. For station $s$, $D_i^{(s)}$ is the detection indicator for event $i$, and
$
X_i^{(s)} = \bigl(X_{i,\mathrm{time}}^{(s)}, X_{i,\mathrm{log(a/T)}}^{(s)}\bigr)
$
denotes the observed summary variables  (arrival time and log(a/T)). Lowercase symbols denote observed values, whereas uppercase symbols denote random variables. The term
$
\pi_s(\theta_i) = P\!\left(D_i^{(s)}=1;\theta_i\right)
$
is the probability that station $s$ detects an event with attributes $\theta_i$. The terms
$
f_{X_{i,\mathrm{time}}^{(s)} \mid D_i^{(s)}=1;\theta_i}
$
and
$
f_{X_{i,\mathrm{log(a/T)}}^{(s)} \mid D_i^{(s)}=1;\theta_i}
$
are the conditional densities of the observed time and log-amplitude-over-period summaries, respectively.

In this formulation, we assume independence between the observed summaries \citep{arora2013netvisa}. Additionally, we use two model equations:
\begin{align}
X_{i,\mathrm{time}}^{(s)}
&=
\theta_{\mathrm{time}} + g_{\mathrm{time}}\!\left(\Delta_{\theta_i}^{(s)},\theta_{{\mathrm{depth}}}\right) + \epsilon_{\mathrm{time},i}^{(s)},
\\
X_{i,\mathrm{log(a/T)}}^{(s)}
&=
\theta_{\mathrm{mag}} - g_{\mathrm{mag}}\!\left(\Delta_{\theta_i}^{(s)},\theta_{\mathrm{depth}} \right) + \epsilon_{\mathrm{mag},i}^{(s)}.
\label{eq:arrival_time_log_at}
\end{align}

\noindent Both functions depend on the distance between the station and the event, $\Delta_{\theta_i}^{(s)}$, as well as the event depth.

Here, \(g_{\mathrm{time}}\) is the model-based seismic travel time \citep{kenntt1991travel}, and \(g_{\mathrm{mag}}\) describes the distance and depth-dependent decay of \(\log(a/T)\) using the same formula applied by the IDC \citep{veith1972magnitude}. We model the remaining errors empirically using three alternatives for robustness, specifically, kernel density estimation (KDE) with an Epanechnikov kernel, a normal distribution, and a Student-\(t\) distribution, with the same error model applied to all stations.

We estimated the detection probability \(\pi_s(\theta_i)\) separately for each station using two models: an expert-guided probit model motivated by a competing (normally distributed) signal-noise argument, and a random forest selected for its validation performance. We then formed decomposed score features for detections, non-detections, \(\log(a/T)\), and arrival time, normalizing each by the number of relevant contributing stations. For \(\log(a/T)\), we followed practice at the IDC of including only observations whose distance was between $20^\circ$ and $100^\circ$. We also included the number of stations contributing to each component (count-based features), event magnitude, and event depth, while deliberately excluding event location as a classification feature. The classifiers we compare are described in Table~\ref{tab:classifiers_ctbt}.

\begin{table}[!htbp]
\centering
\caption{Classifiers considered in the CTBT monitoring example. 
\textit{Baseline} denotes the auxiliary features: event depth, event magnitude, and the count-based features. \textit{Decomposed} denotes the normalized model-fit components obtained by separating detections and non-detections, $\log(a/T)$ and arrival time components. \textit{Observed data} denotes only the $\log(a/T)$ and arrival time part of the decomposed representation. 
\textit{Raw patterns} are station-level indicators of whether a station was active and whether it detected the event.}
\label{tab:classifiers_ctbt}
\small
\begin{tabular}{lll}
\toprule
\textbf{Name} & \textbf{Classifier} & \textbf{Input representation} \\
\midrule
LR-decomp & Logistic regression & Baseline + decomposed \\

LR-obs & Logistic regression & Baseline + observed data \\

LR-baseline & Logistic regression & Baseline \\

RF-raw & Random forest & Raw patterns \\

RF-raw+features & Random forest & Raw patterns + Baseline + decomposed  \\

\bottomrule
\end{tabular}
\end{table}

\subsection{Results}

\subsubsection{Classification}

We evaluate classification performance using the area under
the ROC curve (AUROC) and the area under the precision--recall
curve (AUPRC) for discrimination, Brier score and log-loss
for probabilistic accuracy, and the true negative rate (TNR)
at a fixed true positive rate (TPR) of \(0.95\), thus
quantifying the ability to reject invalid events while
missing at most \(5\%\) of valid events. 

\begin{figure}[!htbp]
\centering
\includegraphics[width=0.5\textwidth]{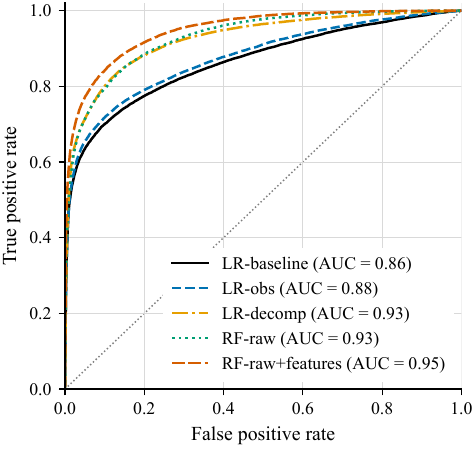}
\caption{ROC curves for all models on the test set. The decomposed feature representation yields a clear improvement over the baseline logistic models, and the best overall performance is obtained by the random forest augmented with the proposed features.}
\label{fig:ROC_main}
\end{figure}

Table~\ref{tab:results_main} summarizes the predictive performance of the classifiers on the test set, and Figure~\ref{fig:ROC_main} shows the corresponding ROC curves. The baseline logistic regression model already provides meaningful
discrimination, with an AUROC of \(0.865\) and an AUPRC of \(0.856\). Adding only the observed-data features leads to a modest improvement (AUROC \(=0.876\), AUPRC \(=0.864\)), indicating that the fit of the observed arrival attributes carries useful signal, but does not by itself capture the full structure of the problem.

A substantially larger improvement is obtained when the feature representation is expanded to include detection score components. Relative to LR-baseline, LR-decomp increases AUROC from \(0.865\) to \(0.925\), improves AUPRC from \(0.856\) to \(0.907\), reduces the Brier score from \(0.131\) to \(0.101\), and reduces log-loss from \(0.423\) to \(0.344\). The gain is especially pronounced at high sensitivity: at a fixed \(\mathrm{TPR}=0.95\), the TNR increases from \(0.302\) to \(0.592\), which is particularly important in our application, where missing valid events is much more costly. This provides strong empirical support for the claim that informative missingness is a central part of the classification problem, and that features designed to capture it are valuable. The ROC curves in Figure~\ref{fig:ROC_main} reinforce this conclusion. The model based on the decomposed representation dominates the simpler logistic baselines over nearly the entire range of false positive rates.

The RF models achieve the best overall performance, which is consistent with the usual trade-off between interpretability and predictive accuracy. LR-decomp performs comparably to RF-raw, despite using a much smaller and more interpretable feature set. This suggests that much of the predictive signal available in the raw station-level inputs is captured by the proposed expert-guided score construction. Moreover, the comparison between RF-raw and RF-raw+features is also informative: augmenting the raw representation with the proposed features increases AUROC from \(0.931\) to \(0.948\), AUPRC from \(0.911\) to \(0.932\), and the TNR at \(\mathrm{TPR}=0.95\) from \(0.642\) to \(0.717\). Hence, the expert-guided features are not only useful for building a transparent classifier, but also remain beneficial when combined with a flexible black-box learner.

\begin{table}[!htbp]
\centering
\caption{Model performance comparison on the test set.}
\label{tab:results_main}
\footnotesize
\begin{tabular}{llllll}
\toprule
\textbf{Model name} & \textbf{AUROC} & \textbf{AUPRC} & \textbf{Brier} & \textbf{Log-Loss} & \textbf{TNR at TPR=0.95}\\
\midrule
LR-baseline       & 0.865 & 0.856 & 0.131 & 0.423 & 0.302 \\
LR-obs            & 0.876 & 0.864 & 0.126 & 0.414 & 0.346 \\
LR-decomp         & 0.925 & 0.907 & 0.101 & 0.344 & 0.592 \\
RF-raw            & 0.931 & 0.911 & 0.100 & 0.325 & 0.642 \\
RF-raw+features   & 0.948 & 0.932 & 0.087 & 0.286 & 0.717 \\
\bottomrule
\end{tabular}
\end{table}

\subsubsection{Expert-guided model}

We briefly assess the expert-guided models directly on the SEL1 dataset. The aim is to examine whether the fitted expert-guided models (constructed by combining prior knowledge with estimation from the separated LEB data), provide a better description of valid SEL1 events, which form the modeled class, than of invalid ones.

\begin{figure}[!htbp]
\centering

\begin{subfigure}[t]{0.49\textwidth}
    \centering
    \includegraphics[width=\textwidth]{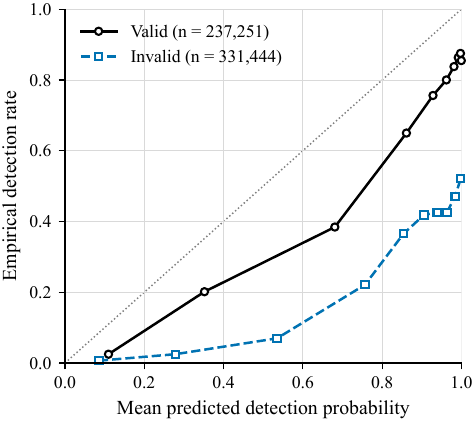}
    \caption{\texttt{WRA}}
    \label{fig:det_station_sel1_WRA}
\end{subfigure}
\hfill
\begin{subfigure}[t]{0.49\textwidth}
    \centering
    \includegraphics[width=\textwidth]{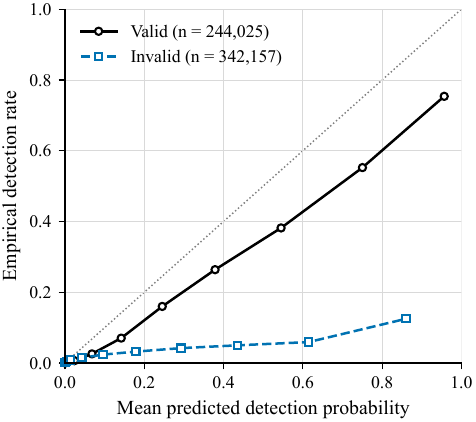}
    \caption{\texttt{BRTR}}
    \label{fig:det_station_sel1_BRTR}
\end{subfigure}

\caption{Calibration plots for the expert-guided station-level detection models on the SEL1 dataset. The left panel (a) shows \texttt{WRA}, and the right panel (b) shows \texttt{BRTR}. In each panel, observed detection frequencies are plotted against predicted detection probabilities for valid and invalid SEL1 events. The dashed diagonal indicates perfect calibration.}
\label{fig:det_station_sel1}

\end{figure}

For the detection mechanism, we examine here two illustrative stations, \texttt{WRA} and \texttt{BRTR}. Figure~\ref{fig:det_station_sel1} and Table~\ref{tab:det_station_sel1} show how the expert-guided probit models fit valid SEL1 events better than invalid ones: valid events are better calibrated, have lower log-loss, and yield higher AUROC. The models also tend to overestimate detection probabilities, likely because SEL1 contains many more non-detections than the LEB data used for fitting.

\begin{table}[!htbp]
\centering
\caption{Performance of the station-level probit detection models on the SEL1 dataset for the two illustrative stations.}
\label{tab:det_station_sel1}
\footnotesize
\begin{tabular}{llccc}
\toprule
\textbf{Station} & \textbf{Subset} & \textbf{AUROC} & \textbf{Brier} & \textbf{Log-loss} \\
\midrule
\texttt{WRA}  & Valid   & 0.827 & 0.176 & 0.721 \\
              & Invalid & 0.747 & 0.383 & 1.475 \\
\texttt{BRTR} & Valid   & 0.889 & 0.122 & 0.417 \\
              & Invalid & 0.765 & 0.143 & 0.462 \\
\bottomrule
\end{tabular}
\end{table}

\begin{figure}[!htbp]
\centering
\begin{subfigure}[t]{0.49\textwidth}
    \centering
    \includegraphics[width=\textwidth]{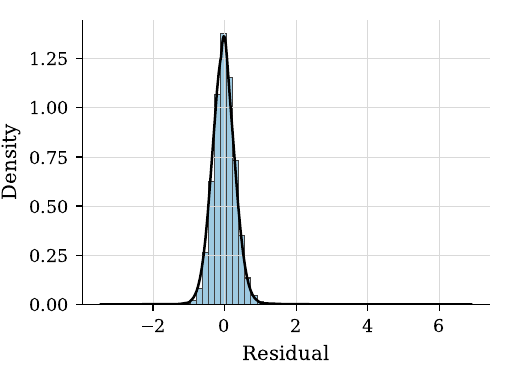}
    \caption{Valid SEL1 \(\log(a/T)\) residuals}
    \label{fig:residual_sel1_logap_valid}
\end{subfigure}
\hfill
\begin{subfigure}[t]{0.49\textwidth}
    \centering
    \includegraphics[width=\textwidth]{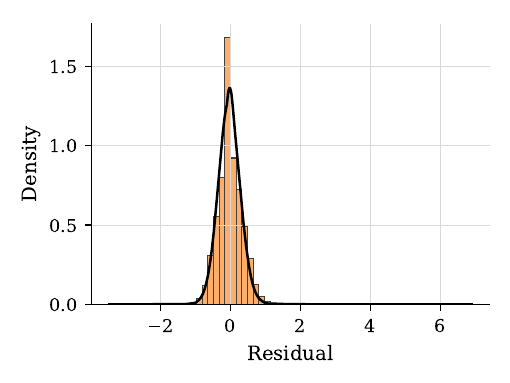}
    \caption{Invalid SEL1 \(\log(a/T)\) residuals}
    \label{fig:residual_sel1_logap_invalid}
\end{subfigure}

\vspace{0.5em}

\begin{subfigure}[t]{0.49\textwidth}
    \centering
    \includegraphics[width=\textwidth]{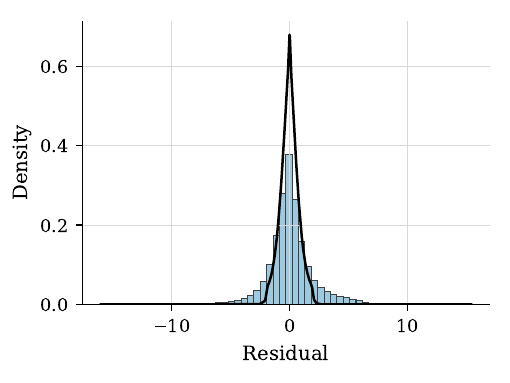}
    \caption{Valid SEL1 arrival-time residuals}
    \label{fig:residual_sel1_time_valid}
\end{subfigure}
\hfill
\begin{subfigure}[t]{0.49\textwidth}
    \centering
    \includegraphics[width=\textwidth]{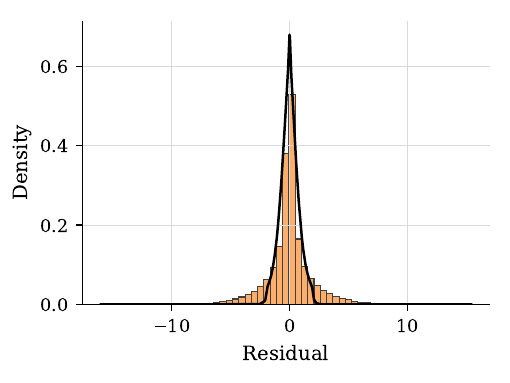}
    \caption{Invalid SEL1 arrival-time residuals}
    \label{fig:residual_sel1_time_invalid}
\end{subfigure}
\caption{Histograms of pooled SEL1 residuals. Solid black curves represent KDEs fitted to the LEB residuals, based on the expert-guided models. The top row shows \(\log(a/T)\) residuals for valid events (a) and invalid events (b); the bottom row shows arrival-time residuals for valid events (c) and invalid events (d).}
\label{fig:residual_fit_sel1}
\end{figure}

For the arrival-time and \(\log(a/T)\) components, we subtract the model-based terms in Equation~\ref{eq:arrival_time_log_at} and pool the residuals across stations and SEL1 events. Table~\ref{tab:residual_fit_sel1} and Figure~\ref{fig:residual_fit_sel1} show that these residual features separate valid and invalid events less strongly than the detection features, especially for arrival time, since associated arrivals must already be broadly time-consistent with the proposed event. The solid black curves in Figure~\ref{fig:residual_fit_sel1} represent Epanechnikov kernel KDEs fitted to the LEB residuals. The residual models fit both classes reasonably well, though arrival-time residuals have somewhat heavier tails, consistent with the need for manual timing adjustments in analyst-reviewed bulletins.

\begin{table}[!htbp]
\centering
\caption{Fit of the pooled residual models on a random subsample of 10{,}000 SEL1 events and all arrivals associated with those events.}
\label{tab:residual_fit_sel1}
\footnotesize
\begin{tabular}{lllc}
\toprule
\textbf{Residual type} & \textbf{Model} & \textbf{Class} & \textbf{Average log predictive density} \\
\midrule
Arrival time & Normal         & Valid   & -3.7977 \\
             &                & Invalid & -3.4974 \\
             & Student-\(t\)  & Valid   & -2.9216 \\
             &                & Invalid & -2.6603 \\
             & KDE            & Valid   & -2.6762 \\
             &                & Invalid & -2.3445 \\
\midrule
\(\log(a/T)\) & Normal        & Valid   & -0.2290 \\
              &               & Invalid & -0.4754 \\
              & Student-\(t\) & Valid   & -0.2238 \\
              &               & Invalid & -0.4074 \\
              & KDE           & Valid   & -0.2230 \\
              &               & Invalid & -0.3989 \\
\bottomrule
\end{tabular}
\end{table}

\section{Simulations}
\label{sec:simulations}

We conducted a simulation study to evaluate the proposed method. The simulation study addresses the following questions:

\begin{enumerate}
    \item Does a classifier based on class-conditional goodness-of-fit score features outperform simple transparent baselines?
    \item Do detection-pattern score components improve performance beyond observed likelihoods alone, particularly as missingness becomes more informative?
    \item Relative to a standard machine learning benchmark (RF) trained on the raw data, how much performance is lost, if any, when using interpretable engineered features with a simple classifier (logistic regression)?
    \item Do the extracted expert-guided features improve performance for the ML benchmark, and does their benefit depend on training sample size?
\end{enumerate}

We also examined two additional aspects: robustness to expert-model misspecification, and the interpretability and stability of the features. The evaluation methods and results for these analyses are provided in the Supplementary Material.

\subsection{Protocol}

The specific setting of false events in the IDC Bulletins cannot be simulated directly, as we lack a generating model for data from these events. Instead, we construct a problem that shares the fundamental structure of the seismic bulletin but allows us to define a mechanism for false events.

We consider binary labels \(Y\in\{0,1\}\), where \(Y=1\) denotes the \emph{valid event} (modeled) class and \(Y=0\) denotes the \emph{invalid event} class. Each instance consists of sensor-level detection indicators and sensor-level measurements observed only upon detection. There are \(S\) sensors indexed by \(s=1,\dots,S\), with sensor locations \(r_s\sim U[0,1]\), drawn once for each simulation replicate. Each event is associated with a latent state $\theta=(L,M)$,
where \(L \sim U[0,1]\) is a location variable and \(M \sim N(10,2^2)\) is a size variable (like the magnitude of a seismic event). In addition, each sensor \(s\) is assigned a noise parameter
$
\alpha_{0s}\sim U[0,1],
$
drawn independently, once at the start of each replicate, and held fixed across instances.

For an instance with \(\theta=(L,M)\), sensor \(s\) detects the event with probability
\begin{equation}
p_s(L,M;\lambda)
=
\sigma\!\left(\alpha_0+\alpha_{0s} + \lambda\bigl(\alpha_M M - \alpha_d |L-r_s|\bigr)\right),
\label{eq:sim_det_model}
\end{equation}
where \(\sigma(\cdot)\) is the logistic function and \(\lambda\) controls the informativeness of non-detects. Any event detected by fewer than two sensors is excluded and replaced by resampling until the desired sample size is reached. If sensor \(s\) detects (\(D^{(s)}=1\)), it provides an observation like amplitude in the seismic application, generated as
\begin{equation}
X^{(s)} \mid \{D^{(s)}=1,\theta,Y=1\}
\sim
\mathcal N\!\left(\beta_0+\beta_M M-\beta_d |L-r_s|,\ \sigma_x^2\right).
\label{eq:sim_obs_model}
\end{equation}
If \(D^{(s)}=0\), this measurement is not observed. 

\paragraph{Invalid events.}
To mimic false seismic events, we consider two types of invalid instances. In the first scenario, an invalid event is created by combining evidence from two sources. Concretely, we draw two latent pseudo-events, $\theta^A=(L^A,M^A)$ and $\theta^B=(L^B,M^B)$,
independently from the valid-event distribution. Then, for each source $s$, we draw an independent selector
$
C^{(s)} \sim \mathrm{Bernoulli}(\gamma),
$
and assign source $s$ to $\theta^A$ or $\theta^B$ accordingly. Sensor detections and observed values are then generated using \eqref{eq:sim_det_model}--\eqref{eq:sim_obs_model}, conditional on the selected pseudo-event. This produces an invalid instance whose measurements are assembled from two incompatible latent events, similar to a falsely created seismic event from associating detections that are close in time but originate from two different events.

In the second scenario, an invalid event has an irregular missingness (detection) pattern. We first draw $(L,M)$ from the same distribution as for valid events. However, detections are then sampled independently across sources, and independently of $(L,M)$ and $r_s$, according to $D^{(s)} \sim \mathrm{Bernoulli}(p_{\mathrm{mal}})$ for $s=1,\dots,S.$
Whenever $D^{(s)}=1$, the corresponding observation $X^{(s)}$ is generated using \eqref{eq:sim_obs_model}. Thus, the observed values remain individually plausible, but the joint detection pattern is inconsistent with the valid generative mechanism. We generate invalid events as a random, 50-50 mixture of these two scenarios. For the valid class, any event detected by fewer than two sources is excluded and replaced by resampling.

\paragraph{Scenario grid.}
The simulation varies two parameters in the main design: the informativeness parameter \(\lambda\), which takes two levels, low (\(1\)) and high (\(2\)), and the training sample size, which takes three levels (\(100\), \(1{,}000\), and \(10{,}000\)). For each value of \(\lambda\), we recalibrate \(\alpha_0\) so that the number of detecting sensors, averaged over all the distributions, remains constant. Events were sampled to obtain exactly 50\% valid and invalid events. All other parameters, which are held fixed across simulations, are reported in the Supplementary Material.

\paragraph{Fitting and features.} An expert-guided model is available only for the modeled class \(Y=1\), which is assumed known up to the instance-specific latent state \(\theta_i=(L_i,M_i)\). In particular, we assume the model parameters \(\alpha_0,\alpha_M,\alpha_d,\beta_0,\beta_M,\beta_d,\sigma_x^2\), the sensor-specific noise parameters and locations (\(\{\alpha_{0s}\}_{s=1}^S,\{r_s\}_{s=1}^S\)), and the scenario parameter \(\lambda\) are known and only \(L_i\) and \(M_i\) are estimated separately for each instance. For each observed instance \(z_i=(\widetilde X_i,D_i)\), we estimate the latent state \(\hat\theta_i=(\hat L_i,\hat M_i)\) by maximizing the class-1 observed-data log-likelihood. Then, at the fitted state \(\hat\theta_i\), we  compute the normalized likelihood features
$
\bar{\ell}_{i,\mathrm{obs}}^{(1)},
\bar{\ell}_{i,\mathrm{det}}^{(1)}, 
\bar{\ell}_{i,\mathrm{nondet}}^{(1)}
$ (see the Supplementary Material for details).

In addition, we include the transparent auxiliary feature vector
$
a_i := \bigl(N_{i,\mathrm{det}},\; \hat M_i,\; \bar R_i,\; s_{R,i}\bigr),
$
where \(N_{i,\mathrm{det}}=\sum_{s=1}^S D_i^{(s)}\) is the number of detecting sources,
$$
R_i^{(s)} := X_i^{(s)} - \bigl(\beta_0+\beta_M \hat M_i-\beta_d |\hat L_i-r_s|\bigr),
\qquad s \text{ such that } D_i^{(s)}=1,
$$
is the residual at source \(s\), \(\bar R_i\) and \(s_{R,i}\) denote, respectively, the sample mean and sample standard deviation of the residuals
$
\{R_i^{(s)}: D_i^{(s)}=1\}.
$

\subsubsection{Methods compared and evaluation}

The classifiers we compare are described in Table~\ref{tab:classifiers}. 

\begin{table}[!htbp]
\centering
\caption{Classifiers considered in the comparison. Here
\(\widetilde X_i^{(s),0}=X_i^{(s)}\) if \(D_i^{(s)}=1\) and
\(\widetilde X_i^{(s),0}=0\) otherwise, that is, missing entries are imputed with 0.}
\label{tab:classifiers}
\small
\begin{tabular}{lll}
\toprule
\textbf{Name} & \textbf{Classifier} & \textbf{Input representation} \\
\midrule
LR-decomp & Logistic regression &
\(\phi_i^{\mathrm{decomp}}=
\bigl(\bar{\ell}_{i,\mathrm{obs}}^{(1)},
\bar{\ell}_{i,\mathrm{det}}^{(1)}, 
\bar{\ell}_{i,\mathrm{nondet}}^{(1)},a_i\bigr)\) \\

LR-obs & Logistic regression &
\(\phi_i^{\mathrm{obs}}=
\bigl(\bar{\ell}_{i,\mathrm{obs}}^{(1)}, a_i\bigr)\) \\

LR-baseline & Logistic regression &
\(\phi_i^{\mathrm{baseline}}=a_i\) \\

RF-raw & Random forest &
\(\displaystyle
x_i^{\mathrm{raw}}=
\bigl(\widetilde X_i^{(1),0},\dots,\widetilde X_i^{(S),0},
D_i^{(1)},\dots,D_i^{(S)}\bigr)
\) \\

RF-raw+features & Random forest &
\(\displaystyle
x_i^{\mathrm{raw+features}}=
\bigl(x_i^{\mathrm{raw}},\bar{\ell}_{i,\mathrm{obs}}^{(1)},
\bar{\ell}_{i,\mathrm{det}}^{(1)}, 
\bar{\ell}_{i,\mathrm{nondet}}^{(1)},a_i\bigr)
\) \\
\bottomrule
\end{tabular}

\end{table}

For each value of \(\lambda\), we generate independent training and test sets and repeat the experiment over \(R\) Monte Carlo replicates. Within each replicate, all methods are evaluated on the same simulated datasets, yielding a paired design. We evaluate predictive performance using the same metrics as in Section~\ref{sec:data}. For each metric and scenario, results are summarized by the Monte Carlo mean and standard error. 

\subsection{Results}

\begin{figure}[!htbp]
\centering

\begin{subfigure}[t]{0.49\textwidth}
    \centering
    \includegraphics[width=\textwidth]{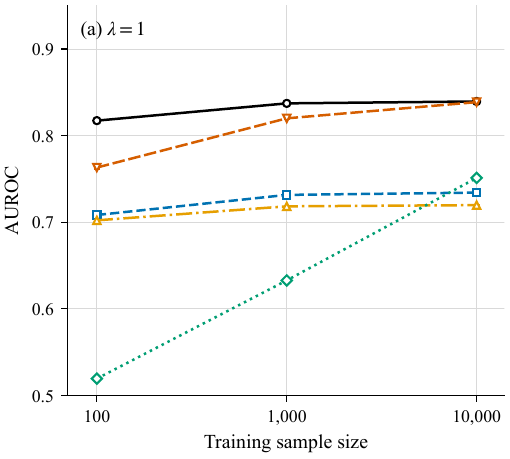}
    \caption{AUROC, \(\lambda=1\)}
    \label{fig:sim_auroc_lam1}
\end{subfigure}
\hfill
\begin{subfigure}[t]{0.49\textwidth}
    \centering
    \includegraphics[width=\textwidth]{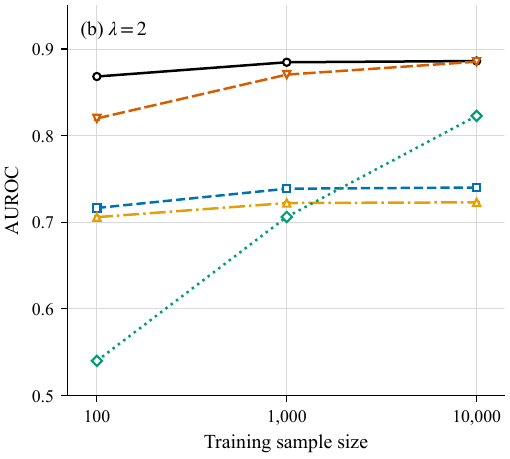}
    \caption{AUROC, \(\lambda=2\)}
    \label{fig:sim_auroc_lam2}
\end{subfigure}

\vspace{0.5em}

\begin{subfigure}[t]{0.49\textwidth}
    \centering
    \includegraphics[width=\textwidth]{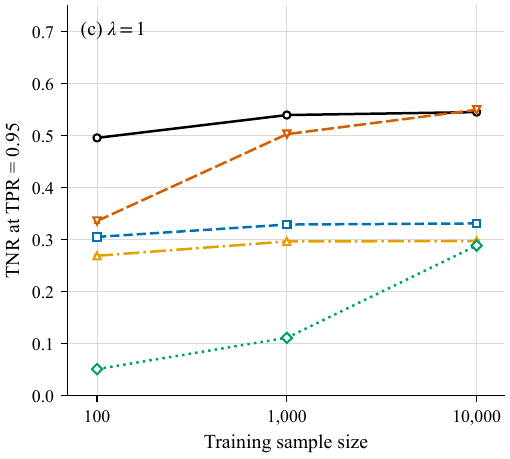}
    \caption{TNR at TPR\(=0.95\), \(\lambda=1\)}
    \label{fig:sim_tnr_lam1}
\end{subfigure}
\hfill
\begin{subfigure}[t]{0.49\textwidth}
    \centering
    \includegraphics[width=\textwidth]{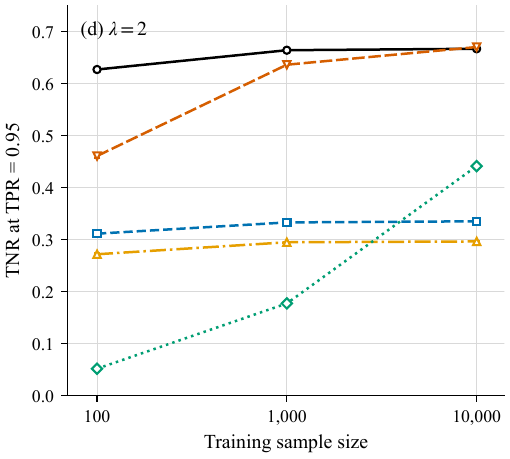}
    \caption{TNR at TPR\(=0.95\), \(\lambda=2\)}
    \label{fig:sim_tnr_lam2}
\end{subfigure}

\vspace{0.7em}

\includegraphics[width=0.80\textwidth]{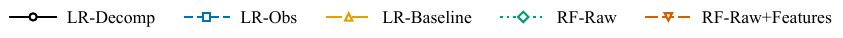}

\caption{Predictive performance across training sample sizes.
The top row shows AUROC and the bottom row shows
TNR at TPR\(=0.95\).
The left panels correspond to the low-\(\lambda\) scenario
(\(\lambda=1\)), and the right panels to the
high-\(\lambda\) scenario (\(\lambda=2\)).
For each method, points denote the Monte Carlo mean
across replicates; Monte Carlo standard errors were below 0.01.}
\label{fig:sim-performance-vs-n}

\end{figure}

Figure~\ref{fig:sim-performance-vs-n} shows how two selected performance metrics change with training sample size under both low- and high-\(\lambda\) scenarios. Detailed results are provided in the Supplementary Material. Overall, Figure~\ref{fig:sim-performance-vs-n} shows a clear ordering among the five methods. LR-decomp is the strongest approach and the best overall across training sample sizes, while RF-raw+features is typically the closest competitor. LR-obs and LR-baseline form a distinctly weaker tier with only a small gap between them. RF-raw performs poorly when \(n\) is small; it improves substantially as \(n\) increases, but still lags clearly behind LR-decomp and RF-raw+features. Thus, a classifier based on class-conditional goodness-of-fit score features substantially outperforms simple transparent baselines. In particular, the gain from LR-baseline to LR-obs is modest, whereas the gain from either of those models to LR-decomp is large and persistent at every \(n\), indicating that the detection-pattern score components contribute predictive information beyond observed likelihoods alone, with this benefit especially evident in the setting where missingness is more informative. 

Relative to the standard nontransparent benchmark, LR-decomp outperforms RF-raw at all sample sizes, although the gap narrows as \(n\) increases, indicating that the transparent feature-engineered approach is most dominant in the small-sample regime. Finally, augmenting the random forest with the extracted expert-guided features improves performance over RF-raw for every \(n\), demonstrating that these features remain useful even for a strong flexible learner: at small sample sizes, RF-raw+features nearly matches LR-decomp, and at sufficiently large \(n\) it even surpasses the other methods.

\section{Conclusions}
\label{sec:conclus}

We proposed an interpretable classification framework for settings with pervasive informative missingness and partial expert knowledge. The central idea is to encode expert knowledge through a class-conditional model under some class label(s), and then convert agreement with that model into a small set of interpretable features. This yields a hybrid procedure in which the model-based component leads to meaningful summaries, while the final prediction is made by a simple discriminative classifier.

In the CTBT monitoring application, this framework leads to a screening rule that mirrors the logic used by human analysts: whether the observed data and the pattern of detections look plausible for a proposed event, given the generative model for data from a true seismic event. The empirical results show that these decomposed model-fit features substantially improve simpler transparent baselines, that explicitly incorporating non-detections is important for performance, and that the features may enhance even standard strong ML classifiers. The simulation study further isolates the contribution of the proposed representation and shows that, particularly in smaller-sample regimes, the interpretable feature-engineered approach can be competitive with, and even outperform, strong standard ML alternatives. The extracted features also remain useful when supplied to a more flexible black-box classifier.

Beyond seismic monitoring, our broader message is that expert knowledge need not be translated into a complete deterministic rule or a complete generative model in order to be useful for classification. Even a partial class-conditional model can provide a principled and interpretable representation of model fit, including the contribution of missingness. This makes the framework relevant to other scientific and operational settings in which observations are naturally evaluated relative to context-dependent expectations. 

More generally, the framework is modular: the model-fit representation can be enriched using multiple expert models, alternative transparent goodness-of-fit scores, uncertainty-aware fitted states, richer notions of sensor availability, hierarchical sensor groupings, or carefully chosen interaction terms. These extensions preserve the central principle of the paper: expert knowledge is used not to impose a rigid decision rule, but to construct interpretable features rooted in statistical theory that expose where and how an observation agrees or disagrees with scientifically meaningful expectations.

\section*{Acknowledgments}
This work was supported by the Pazy Foundation under Grant 396.

\section*{Disclosure statement}
The authors report there are no competing interests to declare.

\section*{Code and data availability}

Simulation code is available at \url{https://github.com/Cohen-Shahar/class-conditional-gof}.

This study used data from the Standard Event List 1 (SEL1) and the Late Event Bulletin (LEB) produced by the International Data Centre (IDC) of the Comprehensive Nuclear-Test Ban Treaty Organization (CTBTO). The data were retrieved from the CTBTO external database in Vienna, Austria. These data are not publicly available and are accessible to CTBTO contractors via the virtual Data Exploitation Center (vDEC).


\bibliographystyle{apalike} 
\bibliography{references} 

@article{rudin2019stop,
  author  = {Rudin, Cynthia},
  title   = {Stop Explaining Black Box Machine Learning Models for High Stakes Decisions and Use Interpretable Models Instead},
  journal = {Nature Machine Intelligence},
  volume  = {1},
  number  = {5},
  pages   = {206--215},
  year    = {2019},
  doi     = {10.1038/s42256-019-0048-x}
}

@inproceedings{caruana2015intelligible,
  author    = {Caruana, Rich and Lou, Yin and Gehrke, Johannes and Koch, Paul and Sturm, Marc and Elhadad, Noemie},
  title     = {Intelligible Models for HealthCare: Predicting Pneumonia Risk and Hospital 30-day Readmission},
  booktitle = {Proceedings of the 21st ACM SIGKDD International Conference on Knowledge Discovery and Data Mining},
  pages     = {1721--1730},
  year      = {2015},
  doi       = {10.1145/2783258.2788613}
}

@book{little2019statistical,
  author    = {Little, Roderick J. A. and Rubin, Donald B.},
  title     = {Statistical Analysis with Missing Data},
  edition   = {3},
  year      = {2019},
  publisher = {John Wiley \& Sons},
  doi       = {10.1002/9781119482260}
}

@inproceedings{lipton2016directly,
  author    = {Lipton, Zachary C. and Kale, David C. and Wetzel, Randall},
  title     = {Directly Modeling Missing Data in Sequences with {RNNs}: Improved Classification of Clinical Time Series},
  booktitle = {Proceedings of the 1st Machine Learning for Healthcare Conference},
  series    = {Proceedings of Machine Learning Research},
  volume    = {56},
  pages     = {253--270},
  year      = {2016},
  publisher = {PMLR},
  url       = {https://proceedings.mlr.press/v56/Lipton16.html}
}

@article{che2018recurrent,
  author  = {Che, Zhengping and Purushotham, Sanjay and Cho, Kyunghyun and Sontag, David and Liu, Yan},
  title   = {Recurrent Neural Networks for Multivariate Time Series with Missing Values},
  journal = {Scientific Reports},
  volume  = {8},
  pages   = {6085},
  year    = {2018},
  doi     = {10.1038/s41598-018-24271-9}
}

@article{karpatne2017theory,
  author  = {Karpatne, Anuj and Atluri, Gowtham and Faghmous, James H. and Steinbach, Michael and Banerjee, Arindam and Ganguly, Auroop R. and Shekhar, Shashi and Samatova, Nagiza and Kumar, Vipin},
  title   = {Theory-Guided Data Science: A New Paradigm for Scientific Discovery from Data},
  journal = {IEEE Transactions on Knowledge and Data Engineering},
  volume  = {29},
  number  = {10},
  pages   = {2318--2331},
  year    = {2017},
  doi     = {10.1109/TKDE.2017.2720168}
}

@article{deng2020integrating,
  author  = {Deng, Changyu and Ji, Xunbi and Rainey, Colton and Zhang, Jianyu and Lu, Wei},
  title   = {Integrating Machine Learning with Human Knowledge},
  journal = {iScience},
  volume  = {23},
  number  = {11},
  pages   = {101656},
  year    = {2020},
  doi     = {10.1016/j.isci.2020.101656}
}

@article{vonrueden2023informed,
  author  = {von Rueden, Laura and Mayer, Sebastian and Beckh, Katharina and Georgiev, Bogdan and Giesselbach, Sven and Heese, Raoul and Kirsch, Birgit and Pfrommer, Julius and Pick, Annika and Ramamurthy, Rajkumar and Walczak, Michal and Garcke, Jochen and Bauckhage, Christian and Schuecker, Jannis},
  title   = {Informed Machine Learning---A Taxonomy and Survey of Integrating Prior Knowledge into Learning Systems},
  journal = {IEEE Transactions on Knowledge and Data Engineering},
  volume  = {35},
  number  = {1},
  pages   = {614--633},
  year    = {2023},
  doi     = {10.1109/TKDE.2021.3079836}
}

@article{richards2009monitoring,
  author  = {Richards, Paul G. and Kim, Won-Young},
  title   = {Monitoring for Nuclear Explosions},
  journal = {Scientific American},
  volume  = {300},
  number  = {3},
  pages   = {70--77},
  year    = {2009},
  doi     = {10.1038/scientificamerican0309-70}
}

@article{draelos2012false,
  author  = {Draelos, Timothy J. and Procopio, Michael J. and Lewis, Jennifer E. and Young, Christopher J.},
  title   = {False Event Screening Using Data Mining in Historical Archives},
  journal = {Seismological Research Letters},
  volume  = {83},
  number  = {2},
  pages   = {267--274},
  year    = {2012},
  doi     = {10.1785/gssrl.83.2.267}
}

@article{benhorin2013scoring,
  author  = {{Ben Horin}, Yochai and Steinberg, David M.},
  title   = {Scoring {IDC} {SEL1} Events for Identifying Misformed Events},
  journal = {Journal of Seismology},
  volume  = {17},
  number  = {4},
  pages   = {1109--1123},
  year    = {2013},
  doi     = {10.1007/s10950-013-9374-3}
}

@article{gaebler2021performance,
  author  = {Gaebler, Peter J. and Ceranna, Lars},
  title   = {Performance of the International Monitoring System Seismic Network Based on Ambient Seismic Noise Measurements},
  journal = {Pure and Applied Geophysics},
  volume  = {178},
  pages   = {2419--2436},
  year    = {2021},
  doi     = {10.1007/s00024-020-02604-y}
}

@article{hutton2006southern,
  author  = {Hutton, Kate and Hauksson, Egill and Clinton, John and Franck, Joseph and Guarino, Anthony and Scheckel, Nick and Given, Doug and Yong, Alan},
  title   = {Southern California Seismic Network Update},
  journal = {Seismological Research Letters},
  volume  = {77},
  number  = {3},
  pages   = {389--395},
  year    = {2006},
  doi     = {10.1785/gssrl.77.3.389}
}

@article{zhou2024comprehensive,
  author  = {Zhou, Youran and Aryal, Sunil and Bouadjenek, Mohamed Reda},
  title   = {Review for Handling Missing Data with Special Missing Mechanism},
  journal = {arXiv preprint arXiv:2404.04905},
  year    = {2024},
  doi     = {10.48550/arXiv.2404.04905}
}

@article{lin2025addressing,
  author  = {Lin, Shanshan and Groenwold, Rolf H. H. and Mehta, Hemalkumar B. and Kim, Ji Soo and Segal, Jodi B.},
  title   = {Addressing Missingness in Predictive Models That Use Electronic Health Record Data},
  journal = {Annals of Internal Medicine},
  volume  = {178},
  number  = {10},
  pages   = {1451--1463},
  year    = {2025},
  doi     = {10.7326/ANNALS-24-01516}
}

@article{draelos2015new,
  author  = {Draelos, Timothy J. and Ballard, Sanford and Young, Christopher J. and Brogan, Ronald},
  title   = {A New Method for Producing Automated Seismic Bulletins: Probabilistic Event Detection, Association, and Location},
  journal = {Bulletin of the Seismological Society of America},
  volume  = {105},
  number  = {5},
  pages   = {2453--2467},
  year    = {2015},
  doi     = {10.1785/0120150099}
}

@article{arora2013netvisa,
  author  = {Arora, Nimar S. and Russell, Stuart and Sudderth, Erik},
  title   = {{NET-VISA}: Network Processing Vertically Integrated Seismic Analysis},
  journal = {Bulletin of the Seismological Society of America},
  volume  = {103},
  number  = {2A},
  pages   = {709--729},
  year    = {2013},
  doi     = {10.1785/0120120107}
}

@article{tibi2019iterative,
  author  = {Tibi, Rigobert and Encarnacao, Andre V. and Ballard, Sanford and Young, Christopher J. and Brogan, Ronald and Sundermier, Amy},
  title   = {The Iterative Processing Framework: A New Paradigm for Automatic Event Building},
  journal = {Bulletin of the Seismological Society of America},
  volume  = {109},
  number  = {6},
  pages   = {2501--2509},
  year    = {2019},
  doi     = {10.1785/0120190093}
}

@article{mcbrearty2019association,
  author  = {McBrearty, Ian W. and Gomberg, Joan and Delorey, Andrew A. and Johnson, Paul A.},
  title   = {Earthquake Arrival Association with Backprojection and Graph Theory},
  journal = {Bulletin of the Seismological Society of America},
  volume  = {109},
  number  = {6},
  pages   = {2510--2531},
  year    = {2019},
  doi     = {10.1785/0120190081}
}

@incollection{wu2021ctbt,
  author    = {Wu, Zhongliang and Richards, Paul G.},
  title     = {Seismology, Monitoring of {CTBT}},
  booktitle = {Encyclopedia of Solid Earth Geophysics},
  editor    = {Gupta, Harsh K.},
  pages     = {1669--1673},
  publisher = {Springer},
  address   = {Cham},
  year      = {2021},
  doi       = {10.1007/978-3-030-58631-7_163}
}

@article{emmanuel2021survey,
  author  = {Emmanuel, Tlamelo and Maupong, Thabiso and Mpoeleng, Dimane and Semong, Thabo and Mphago, Banyatsang and Tabona, Oteng},
  title   = {A Survey on Missing Data in Machine Learning},
  journal = {Journal of Big Data},
  volume  = {8},
  number  = {1},
  pages   = {140},
  year    = {2021},
  doi     = {10.1186/s40537-021-00516-9}
}

@article{miao2023experimental,
  author  = {Miao, Xiaoye and Wu, Yangyang and Chen, Lu and Gao, Yunjun and Yin, Jianwei},
  title   = {An Experimental Survey of Missing Data Imputation Algorithms},
  journal = {IEEE Transactions on Knowledge and Data Engineering},
  volume  = {35},
  number  = {7},
  pages   = {6630--6650},
  year    = {2023},
  doi     = {10.1109/TKDE.2022.3186498}
}

@article{dwivedi2023xai,
  author     = {Dwivedi, Rudresh and Dave, Devam and Naik, Het and Singhal, Smiti and Omer, Rana and Patel, Pankesh and Qian, Bin and Wen, Zhenyu and Shah, Tejal and Morgan, Graham and Ranjan, Rajiv},
  title      = {Explainable {AI} ({XAI}): Core Ideas, Techniques, and Solutions},
  journal    = {ACM Computing Surveys},
  volume     = {55},
  number     = {9},
  articleno  = {194},
  numpages   = {33},
  year       = {2023},
  doi        = {10.1145/3561048}
}

@article{allen2024interpretable,
  author  = {Allen, Genevera I. and Gan, Luqin and Zheng, Lili},
  title   = {Interpretable Machine Learning for Discovery: Statistical Challenges and Opportunities},
  journal = {Annual Review of Statistics and Its Application},
  volume  = {11},
  pages   = {97--121},
  year    = {2024},
  doi     = {10.1146/annurev-statistics-040120-030919}
}

@article{willard2022integrating,
  author  = {Willard, Jared and Jia, Xiaowei and Xu, Shaoming and Steinbach, Michael and Kumar, Vipin},
  title   = {Integrating Scientific Knowledge with Machine Learning for Engineering and Environmental Systems},
  journal = {ACM Computing Surveys},
  volume  = {55},
  number  = {4},
  pages   = {1--37},
  year    = {2022},
  doi     = {10.1145/3514228}
}

@inproceedings{ng2002discriminative,
  author    = {Ng, Andrew Y. and Jordan, Michael I.},
  title     = {On Discriminative vs. Generative Classifiers: A Comparison of Logistic Regression and Naive Bayes},
  booktitle = {Advances in Neural Information Processing Systems 14},
  pages     = {841--848},
  year      = {2002},
  url       = {https://papers.nips.cc/paper/2020-on-discriminative-vs-generative-classifiers-a-comparison-of-logistic-regression-and-naive-bayes}
}

@book{jebara2004machine,
  author    = {Jebara, Tony},
  title     = {Machine Learning: Discriminative and Generative},
  publisher = {Springer},
  address   = {New York},
  year      = {2004},
  doi       = {10.1007/978-1-4419-9011-2}
}

@inproceedings{raina2003classification,
  author    = {Raina, Rajat and Shen, Yirong and McCallum, Andrew and Ng, Andrew Y.},
  title     = {Classification with Hybrid Generative/Discriminative Models},
  booktitle = {Advances in Neural Information Processing Systems 16},
  year      = {2003},
  url       = {https://papers.nips.cc/paper/2405-classification-with-hybrid-generativediscriminative-models}
}

@inproceedings{jaakkola1998exploiting,
  author    = {Jaakkola, Tommi S. and Haussler, David},
  title     = {Exploiting Generative Models in Discriminative Classifiers},
  booktitle = {Advances in Neural Information Processing Systems 11},
  year      = {1998},
  url       = {https://proceedings.neurips.cc/paper_files/paper/1998/hash/db1915052d15f7815c8b88e879465a1e-Abstract.html}
}

@inproceedings{kuleshov2017deep,
  author    = {Kuleshov, Volodymyr and Ermon, Stefano},
  title     = {Deep Hybrid Models: Bridging Discriminative and Generative Approaches},
  booktitle = {Proceedings of the 33rd Conference on Uncertainty in Artificial Intelligence},
  year      = {2017},
  url       = {https://www.cs.cornell.edu/~kuleshov/papers/uai2017.pdf}
}

@article{hayashi2024hybrid,
  author  = {Hayashi, Hideaki},
  title   = {A Hybrid of Generative and Discriminative Models Based on the Gaussian-Coupled Softmax Layer},
  journal = {IEEE Transactions on Neural Networks and Learning Systems},
  volume  = {36},
  number  = {2},
  pages   = {2894--2904},
  year    = {2025},
  doi     = {10.1109/TNNLS.2024.3358113}
}

@article{brodersen2011generative,
  author  = {Brodersen, Kay H. and Schofield, Thomas M. and Leff, Alexander P. and Ong, Cheng Soon and Lomakina, Ekaterina I. and Buhmann, Joachim M. and Stephan, Klaas E.},
  title   = {Generative Embedding for Model-Based Classification of fMRI Data},
  journal = {PLoS Computational Biology},
  volume  = {7},
  number  = {6},
  pages   = {e1002079},
  year    = {2011},
  doi     = {10.1371/journal.pcbi.1002079}
}

@article{bicego2009component,
  author  = {Bicego, Manuele and P{\k{e}}kalska, El{\.z}bieta and Tax, David M. J. and Duin, Robert P. W.},
  title   = {Component-Based Discriminative Classification for Hidden Markov Models},
  journal = {Pattern Recognition},
  volume  = {42},
  number  = {11},
  pages   = {2637--2648},
  year    = {2009},
  doi     = {10.1016/j.patcog.2009.03.023}
}

@article{perina2012free,
  author  = {Perina, Alessandro and Cristani, Marco and Castellani, Umberto and Murino, Vittorio and Jojic, Nebojsa},
  title   = {Free Energy Score Spaces: Using Generative Information in Discriminative Classifiers},
  journal = {IEEE Transactions on Pattern Analysis and Machine Intelligence},
  volume  = {34},
  number  = {7},
  pages   = {1249--1262},
  year    = {2012},
  doi     = {10.1109/TPAMI.2011.241}
}

@inproceedings{ipsen2022missing,
  author    = {Ipsen, Niels Bruun and Mattei, Pierre-Alexandre and Frellsen, Jes},
  title     = {How to Deal with Missing Data in Supervised Deep Learning?},
  booktitle = {International Conference on Learning Representations},
  year      = {2022},
  url       = {https://openreview.net/forum?id=jEXxzPUMYVZ}
}

@article{pingi2024joint,
  author  = {Pingi, Sharon Torao and Zhang, Duoyi and Bashar, Md. Abul and Nayak, Richi},
  title   = {Joint Representation Learning with Generative Adversarial Imputation Network for Improved Classification of Longitudinal Data},
  journal = {Data Science and Engineering},
  volume  = {9},
  number  = {1},
  pages   = {5--25},
  year    = {2024},
  doi     = {10.1007/s41019-023-00232-9}
}

@article{dimarino2025antehoc,
  author     = {Di Marino, Antonio and Bevilacqua, Vincenzo and Ciaramella, Angelo and De Falco, Ivanoe and Sannino, Giovanna},
  title      = {Ante-Hoc Methods for Interpretable Deep Models: A Survey},
  journal    = {ACM Computing Surveys},
  volume     = {57},
  number     = {10},
  articleno  = {262},
  pages      = {1--36},
  year       = {2025},
  doi        = {10.1145/3728637}
}

@inproceedings{koh2020concept,
  author    = {Koh, Pang Wei and Nguyen, Thao and Tang, Yew Siang and Mussmann, Stephen and Pierson, Emma and Kim, Been and Liang, Percy},
  title     = {Concept Bottleneck Models},
  booktitle = {Proceedings of the 37th International Conference on Machine Learning},
  series    = {Proceedings of Machine Learning Research},
  volume    = {119},
  pages     = {5338--5348},
  year      = {2020},
  url       = {https://proceedings.mlr.press/v119/koh20a.html}
}

@inproceedings{alvarezmellis2018self,
  author    = {Alvarez-Melis, David and Jaakkola, Tommi S.},
  title     = {Towards Robust Interpretability with Self-Explaining Neural Networks},
  booktitle = {Advances in Neural Information Processing Systems 31},
  pages     = {7786--7795},
  year      = {2018},
  url       = {https://papers.nips.cc/paper/8003-towards-robust-interpretability-with-self-explaining-neural-networks.pdf}
}

@article{belenguerllorens2026interpretable,
  author  = {Belenguer-Llorens, Albert and Sevilla-Salcedo, Carlos and Mourao-Miranda, Janaina and G{\'o}mez-Verdejo, Vanessa},
  title   = {Interpretable Generative and Discriminative Learning for Multimodal and Incomplete Clinical Data},
  journal = {Information Fusion},
  volume  = {129},
  pages   = {104070},
  year    = {2026},
  doi     = {10.1016/j.inffus.2025.104070}
}

@inproceedings{mattei2019miwae,
  author    = {Mattei, Pierre-Alexandre and Frellsen, Jes},
  title     = {{MIWAE}: Deep Generative Modelling and Imputation of Incomplete Data Sets},
  booktitle = {Proceedings of the 36th International Conference on Machine Learning},
  series    = {Proceedings of Machine Learning Research},
  volume    = {97},
  pages     = {4413--4423},
  year      = {2019},
  editor    = {Chaudhuri, Kamalika and Salakhutdinov, Ruslan},
  publisher = {PMLR},
  url       = {https://proceedings.mlr.press/v97/mattei19a.html}
}

@inproceedings{ma2020vaem,
  author    = {Ma, Chao and Tschiatschek, Sebastian and Turner, Richard and Hern{\'a}ndez-Lobato, Jos{\'e} Miguel and Zhang, Cheng},
  title     = {{VAEM}: a Deep Generative Model for Heterogeneous Mixed Type Data},
  booktitle = {Advances in Neural Information Processing Systems 33},
  pages     = {11237--11247},
  year      = {2020},
  url       = {https://proceedings.neurips.cc/paper_files/paper/2020/hash/8171ac2c5544a5cb54ac0f38bf477af4-Abstract.html}
}

@inproceedings{gong2021vsae,
  author    = {Gong, Yu and Hajimirsadeghi, Hossein and He, Jiawei and Durand, Thibaut and Mori, Greg},
  title     = {Variational Selective Autoencoder: Learning from Partially-Observed Heterogeneous Data},
  booktitle = {Proceedings of the 24th International Conference on Artificial Intelligence and Statistics},
  series    = {Proceedings of Machine Learning Research},
  volume    = {130},
  pages     = {2377--2385},
  year      = {2021},
  editor    = {Banerjee, Arindam and Fukumizu, Kenji},
  publisher = {PMLR},
  url       = {https://proceedings.mlr.press/v130/gong21a.html}
}

@article{kenntt1991travel,
  author  = {Kennett, B. L. N. and Engdahl, E. R.},
  title   = {Travel Times for Global Earthquake Location and Phase Association},
  journal = {Geophysical Journal International},
  volume  = {105},
  pages   = {429--465},
  year    = {1991},
  doi     = {10.1111/j.1365-246X.1991.tb06724.x}
}

@article{veith1972magnitude,
  author  = {Veith, K. F. and Clawson, G. E.},
  title   = {Magnitude from Short-Period {P}-Wave Data},
  journal = {Bulletin of the Seismological Society of America},
  volume  = {62},
  number  = {2},
  pages   = {435--452},
  year    = {1972},
  doi     = {10.1785/BSSA0620020435}
}

@article{scholkopf2001estimating,
  author  = {Sch{\"o}lkopf, Bernhard and Platt, John C. and Shawe-Taylor, John and Smola, Alex J. and Williamson, Robert C.},
  title   = {Estimating the Support of a High-Dimensional Distribution},
  journal = {Neural Computation},
  volume  = {13},
  number  = {7},
  pages   = {1443--1471},
  year    = {2001},
  doi     = {10.1162/089976601750264965}
}

@article{tax2004support,
  author  = {Tax, David M. J. and Duin, Robert P. W.},
  title   = {Support Vector Data Description},
  journal = {Machine Learning},
  volume  = {54},
  number  = {1},
  pages   = {45--66},
  year    = {2004},
  doi     = {10.1023/B:MACH.0000008084.60811.49}
}

@article{pimentel2014review,
  author  = {Pimentel, Marco A. F. and Clifton, David A. and Clifton, Lei and Tarassenko, Lionel},
  title   = {A Review of Novelty Detection},
  journal = {Signal Processing},
  volume  = {99},
  pages   = {215--249},
  year    = {2014},
  doi     = {10.1016/j.sigpro.2013.12.026}
}

@article{saar2007handling,
  author  = {Saar-Tsechansky, Maytal and Provost, Foster},
  title   = {Handling Missing Values When Applying Classification Models},
  journal = {Journal of Machine Learning Research},
  volume  = {8},
  pages   = {1623--1657},
  year    = {2007},
  url     = {https://www.jmlr.org/papers/v8/saar-tsechansky07a.html}
}

@incollection{Nutovitz2023XAI,
  author    = {Notovich, Aviv and Chalutz-Ben Gal, Hila and Ben-Gal, Irad},
  title     = {Explainable Artificial Intelligence ({XAI}): Motivation, Terminology, and Taxonomy},
  booktitle = {Machine Learning for Data Science Handbook: Data Mining and Knowledge Discovery Handbook},
  pages     = {971--985},
  address   = {Cham},
  publisher = {Springer International Publishing},
  year      = {2023},
  doi       = {10.1007/978-3-031-24628-9_41}
}

@article{karniadakis2021,
  author  = {Karniadakis, George Em and Kevrekidis, Ioannis G. and Lu, Lu and Perdikaris, Paris and Wang, Sifan and Yang, Liu},
  title   = {Physics-Informed Machine Learning},
  journal = {Nature Reviews Physics},
  volume  = {3},
  number  = {6},
  pages   = {422--440},
  year    = {2021},
  doi     = {10.1038/s42254-021-00314-5}
}

@article{schulz2011automatic,
  title={Automatic, three-segment, MR-based attenuation correction for whole-body PET/MR data},
  author={Schulz, V and Torres-Espallardo, I and Renisch, S and Hu, Z and Ojha, N and B{\"o}rnert, P and Perkuhn, M and Niendorf, T and Sch{\"a}fer, WM and Brockmann, H and others},
  journal={European journal of nuclear medicine and molecular imaging},
  volume={38},
  number={1},
  pages={138--152},
  year={2011},
  publisher={Springer}
}

@article{cohen2026improved,
  title={Improved Automatic Seismic Bulletins Via Likelihood-Based Model Fit Scores for Classification},
  author={Cohen, Shahar and Steinberg, David M and Radzyner, Yael and Ben-Horin, Yochai},
  journal={Pure and Applied Geophysics},
  pages={1--31},
  year={2026},
  publisher={Springer}
}

\end{document}